\documentclass[letterpaper, 10 pt, conference]{ieeeconf}  %

\IEEEoverridecommandlockouts                              %

\usepackage{graphicx}
\usepackage{amsmath}
\usepackage{bbm}
\usepackage{dsfont}
\usepackage{subcaption}

\PassOptionsToPackage{dvipsnames}{xcolor}
\PassOptionsToPackage{table}{xcolor}

\usepackage{amssymb}
\usepackage{booktabs}
\usepackage{makecell}
\usepackage[ruled,noend,linesnumbered]{algorithm2e}
\usepackage{titlesec}
\usepackage{wrapfig}
\usepackage{array}
\usepackage{multirow}
\usepackage{colortbl}
\usepackage{tabularx}

\usepackage{grffile}
\usepackage{xspace}
\usepackage{needspace}

\usepackage{comment} 
\usepackage{url}
\usepackage{bm}
\usepackage{paralist}

\usepackage{pifont}

\usepackage[belowskip=0pt,aboveskip=5pt,font=small]{caption}
\usepackage[belowskip=0pt,aboveskip=0pt,font=small]{subcaption}
\usepackage{xcolor}
\PassOptionsToPackage{hyphens}{url}
\usepackage[hidelinks]{hyperref}
\usepackage{xfrac}
\usepackage{cite}

\def\@fnsymbol#1{\ensuremath{\ifcase#1\or *\or \dagger\or \ddagger\or
   \mathsection\or \mathparagraph\or \|\or **\or \dagger\dagger
   \or \ddagger\ddagger \else\@ctrerr\fi}}
\newcommand{\ssymbol}[1]{^{\@fnsymbol{#1}}}

\newcommand{\approachname}{VLFM}
\newcommand{\fullapproachname}{Vision-Language Frontier Maps}
\newcommand{\xhdr}[1]{\vspace{2pt}\textbf{#1}}
\newcommand{\moveforward}{\textsc{move\_forward}\xspace}

\newcommand{\turnleft}{\textsc{turn\_left}\xspace}
\newcommand{\turnright}{\textsc{turn\_right}\xspace}
\newcommand{\stopac}{\textsc{stop}\xspace}
\newcommand{\lookup}{\textsc{look\_up}\xspace}
\newcommand{\lookdown}{\textsc{look\_down}\xspace}
\newcommand{\projecturl}{\href{http://naoki.io/vlfm}{\redtext{naoki.io/vlfm}}}

\newcommand{\targetprompt}{Seems like there is a <target\_object> ahead.}

\newcommand{\redtext}[1]{\textcolor{red}{#1}}

\def\eg{\emph{e.g.}}

\title{\LARGE \bf
\approachname: Vision-Language Frontier Maps \\ for Zero-Shot Semantic Navigation
}

\author{
    Naoki Yokoyama$^{1,2}$,
    Sehoon Ha$^{2}$,
    Dhruv Batra$^{2}$,
    Jiuguang Wang$^{1}$,
    Bernadette Bucher$^{1}$
}

\begin{document}

\twocolumn[{%
  \renewcommand\twocolumn[1][]{#1}%
  \maketitle
  \centering
  \includegraphics[width=0.99\textwidth]{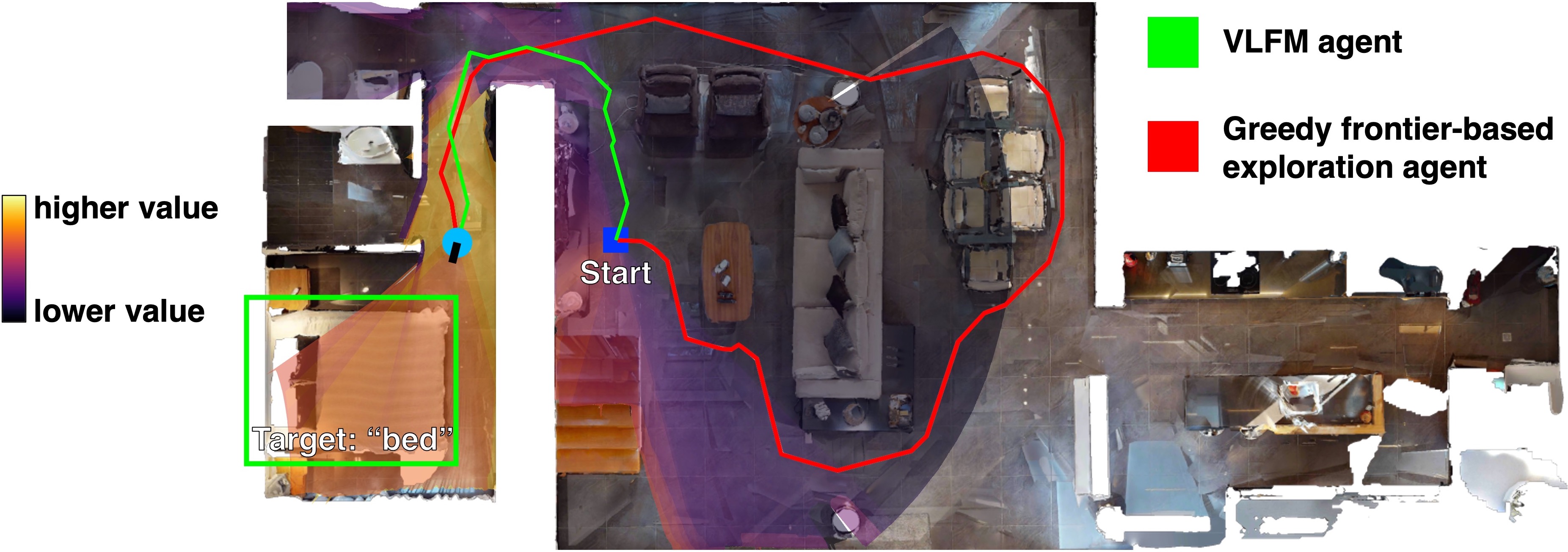}
  \label{fig:teaser}
  \captionof{figure}{\approachname\ achieves state-of-the-art semantic Object Goal Navigation performance in unfamiliar environments, without task-specific training, pre-built maps, or prior knowledge of the surroundings. It utilizes a vision-language model to explore the environment, capitalizing on visual semantic cues that are likely to guide the agent towards the goal to explore the environment more efficiently than a greedy frontier-based exploration agent.}
  
}]
\footnotetext[1]{Work done during NY's internship. JW and BB are with the Boston Dynamics AI Institute {\tt\small \{jw,bbucher\}@theaiinstitute.com}}
\footnotetext[2]{NY, SH, and DB are with the Georgia Institute of Technology {\tt\small \{nyokoyama,sehoonha,dbatra\}@gatech.edu}}%

\thispagestyle{empty}
\pagestyle{empty}

\begin{abstract}
Understanding how humans leverage semantic knowledge to navigate unfamiliar environments and decide where to explore next is pivotal for developing robots capable of human-like search behaviors.
We introduce a zero-shot navigation approach, \fullapproachname\ (\approachname), which is inspired by human reasoning and designed to navigate towards unseen semantic objects in novel environments.
\approachname\ builds occupancy maps from depth observations to identify frontiers, and leverages RGB observations and a pre-trained vision-language model to generate a \emph{language-grounded value map}.
\approachname\ then uses this map to identify the most promising frontier to explore for finding an instance of a given target object category.
We evaluate \approachname\ in photo-realistic environments from the Gibson, Habitat-Matterport 3D (HM3D), and Matterport 3D (MP3D) datasets within the Habitat simulator.
Remarkably, \approachname\ achieves state-of-the-art results on all three datasets as measured by success weighted by path length (SPL) for the Object Goal Navigation task.
Furthermore, we show that \approachname's zero-shot nature enables it to be readily deployed on real-world robots such as the Boston Dynamics Spot mobile manipulation platform.
We deploy \approachname\ on Spot and demonstrate its capability to efficiently navigate to target objects within an office building in the real world, without any prior knowledge of the environment.
The accomplishments of \approachname\ underscore the promising potential of vision-language models in advancing the field of semantic navigation. Videos of real world deployment can be viewed at \projecturl.
\end{abstract}

\section{Introduction}

\noindent How do humans navigate in novel environments? 
The process of human navigation in unfamiliar environments is complex, often relying on a combination of explicit maps and internal knowledge.
This internal knowledge is typically an accumulation of semantic knowledge, which can be used to infer the layout of the space, including the locations of specific objects and geometric configurations.
For instance, we know that toilets and showers are usually found together in bathrooms, often located near bedrooms.
Natural language can further enhance this prior semantic knowledge, depending on the context.

In the development of robots capable of human-like navigation, learned foundation models that mimic this human reasoning process can be invaluable.
Many methods, known as \emph{zero-shot} methods, utilize these models to facilitate semantic navigation without any task-specific training or fine-tuning.
Zero-shot methods are convenient because they can be easily adapted or repurposed for future robotic systems performing complex tasks, and they provide intermediate representations that improve interpretability.
The remarkable performance of large language models (LLMs) and vision-language models (VLMs) has facilitated task-independent solutions for the semantic inference of out-of-view scene information~\cite{gadre2022cow,zhou2023esc,chen2023not}.

In this work, we propose \fullapproachname\ (\approachname), a zero-shot approach for target-driven semantic navigation to an unseen object in a novel environment.
\approachname\ builds occupancy maps from depth observations to identify frontiers of the explored map region.
To find semantic target objects, \approachname\ prompts a pre-trained VLM to select which of these frontiers is most likely to lead to the semantic target.
In contrast to prior language-based zero-shot semantic navigation methods~\cite{zhou2023esc,dorbala2023can,chen2023not}, our method does not rely on object detectors and language models (\eg, ChatGPT, BERT) to evaluate frontiers using text-only semantic reasoning.
Instead, \approachname\ uses a vision-language model to directly extract semantic values from RGB images in the form of a cosine similarity score with a text prompt involving the target object.
\approachname\ uses these scores to generate a \emph{language-grounded value map} that is used to identify the most promising frontier to explore.
This spatially-grounded joint vision-language-based semantic reasoning increases computational inference speed and overall semantic navigation performance.

We demonstrate \approachname\ in photorealistic environments within the Habitat~\cite{habitat19iccv} simulator, where we achieve state-of-the-art results on the Object Goal Navigation (ObjectNav) task, even when compared to methods trained directly on the task.
Specifically, we achieve absolute increases in success rates weighted by path length over prior state-of-the-art approaches of 12\% on Gibson~\cite{xiazamirhe2018gibsonenv}, 5\% on Matterport 3D (MP3D)~\cite{Matterport3D} and 3\% on Habitat-Matterport 3D (HM3D)~\cite{ramakrishnan2021hm3d} datasets.
We also demonstrate our approach in the real world on a Boston Dynamics Spot mobile manipulation platform by navigating efficiently to unseen semantic targets across a novel office building floor, without access to a pre-built map.

\section{Related Works}

\begin{figure*}[t!]
\centering
    \vskip5pt
    \includegraphics[width=1.0\textwidth]{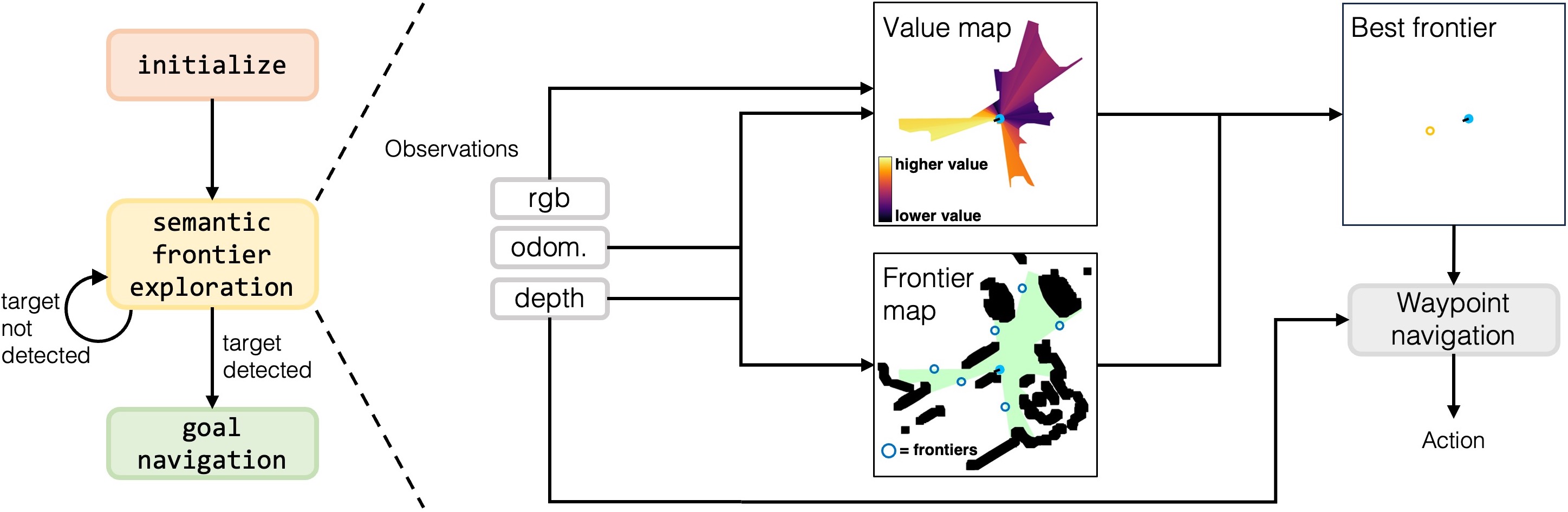}
    \caption{
        \approachname\ constructs an occupancy map of the scene identifying frontiers of explored space as well as a value map of the likelihood of each region to lead toward the out-of-view target object.
        In a navigation episode, the robot first spins in a circle to initialize these maps and then begins executing frontier-based exploration by selecting waypoints from the current map frontier using the value map.
        Navigation to each waypoint is executed with a PointNav policy trained with Variable Experience Rollout (VER)~\cite{wijmans2022ver}.
        The policy is also used to navigate to the target object once it is detected (`goal navigation').
    }
    \label{fig:overview}
    \vspace{-0.2cm}
\end{figure*}

\xhdr{ObjectNav.}
Object Goal Navigation (ObjectNav) involves executing semantic target-driven navigation in a novel environment, where performance is primarily measured by the efficiency of the robot's path to an instance of a given target object category.
This is based on the premise that effective use of semantic priors should enable a robot to locate objects more efficiently~\cite{batra2020objectnav}.
Learning approaches to train robots with semantic navigation abilities have typically leveraged reinforcement learning~\cite{ye2021auxiliary,chang2020semantic,procthor,gireesh22objQlearn}, learning from demonstration~\cite{ramrakhya2023pirlnav}, or prediction of semantic top-down maps~\cite{chaplot2020object, zhang2021hierarchical,luo2022stubborn, ramakrishnan2022poni, zhang20223d,chen2023not} on which waypoint planners can be used.

However, these task-specific trained approaches only work with the closed-set of object categories that they were trained on, and are often trained exclusively on simulated data, which can impede deployment of these policies onto real-world platforms.
In contrast, our work proposes a zero-shot method that can take in an open-set of object categories, uses models that were trained on large amounts of real-world data, and demonstrates successful semantic navigation in the real world.

\xhdr{Zero-shot ObjectNav.}
Recent works in zero-shot methods for ObjectNav involve adapting the frontier-based exploration method proposed by~\cite{yamauchi1997frontier}.
Frontier-based exploration involves visiting the boundaries between explored and unexplored areas on a map that is iteratively built by the agent as it explores.
Many methods for choosing the next frontier to explore have been proposed, such as classical methods that select frontiers based on the expected amount of information the agent would gain~\cite{histogramfbe,li2022learning}. 
CLIP on Wheels (CoW)~\cite{gadre2022cow} adopts a straightforward approach in which the robot explores the closest frontier until the target object is detected using either CLIP~\cite{radford2021learning} features or an open-vocabulary object detector.
LGX~\cite{dorbala2023can} and ESC~\cite{zhou2023esc} use a large language model (LLM) that processes object detections presented in the form of text to identify which frontiers would most likely harbor an instance of the target object.
Instead of an LLM, SemUtil~\cite{chen2023not} uses BERT~\cite{devlin2019bert} to embed the class labels of objects detected near frontiers, and then compare them to the text embedding of the target object to select the frontier to explore next.
However, these methods introduce a bottleneck in which visual cues from the environment must be converted into text by an object detector before they can be used to semantically evaluate frontiers.
Additionally, reliance on an LLM requires a large amount of compute that may require a remote server the robot must connect to.
In contrast, \approachname\ uses a vision-language model that can be easily loaded onto a consumer laptop to generate semantic value scores directly from RGB observations and text prompts, without generating any text from visual observations.

\section{Problem Formulation}

We address the task of ObjectNav~\cite{batra2020objectnav}, where a robot is tasked with searching for an instance of a target object category (\emph{e.g.}, `bed') in a previously unseen environment.
This semantic navigation task encourages the robot to understand and navigate the environment based on high-level semantic concepts, such as the type of room it's in or the types of objects it sees, rather than relying solely on geometric cues. 
The robot only has access to an egocentric \mbox{RGB-D} camera and an odometry sensor that provides its current forward and horizontal distance and heading relative to its starting pose.
The action space consists of the following:
\moveforward ($0.25m$),
\turnleft ($30^{\circ}$),
\turnright ($30^{\circ}$),
\lookup ($30^{\circ}$),
\lookdown ($30^{\circ}$),
and \stopac.
An episode is defined as successfully completed if \stopac is called within 1 m of any instance of the target object in 500 or fewer steps. 

\section{Vision-Language Frontier Maps}
As depicted in Fig. \ref{fig:overview}, our approach is divided into three phases: initialization, exploration, and goal navigation.
In the \emph{initialization} phase, the robot rotates in place for a complete turn to set up its frontier and value maps, which are crucial for the subsequent exploration phase.
During \emph{exploration}, the robot persistently updates the frontier and value maps to create frontier waypoints and select the most valuable one for locating the specified target object category and navigating to it.
Once it detects a target object instance, it transitions to the goal navigation phase.
In the \emph{goal navigation} phase, the robot simply navigates to the nearest point on the detected target object and triggers \stopac once it is within sufficient proximity.

\subsection{Frontier waypoint generation}

We utilize depth and odometry observations to build a top-down 2D map of obstacles that the robot has encountered.
The explored area within this map is updated based on the robot's location, its current heading, and any obstacles that obstruct parts of its current view from being explored.
To identify obstacle locations, we transform the current depth image into a point cloud, filter out any points that are either too short or too tall to be considered an obstacle, transform the points to the global frame, and then project them onto a 2D grid.
We then identify each boundary separating the explored and unexplored areas, identifying its midpoint as a potential frontier waypoint.
As the robot explores the area, the quantity and locations of frontiers will vary until the entire environment has been explored and no more frontiers remain.
If the robot has not detected a target object at this point, it will simply trigger \stopac to end the episode (unsuccessfully).

\subsection{Value map generation}
\begin{figure}[t]
    \centerline{\includegraphics[width=1.0\columnwidth]{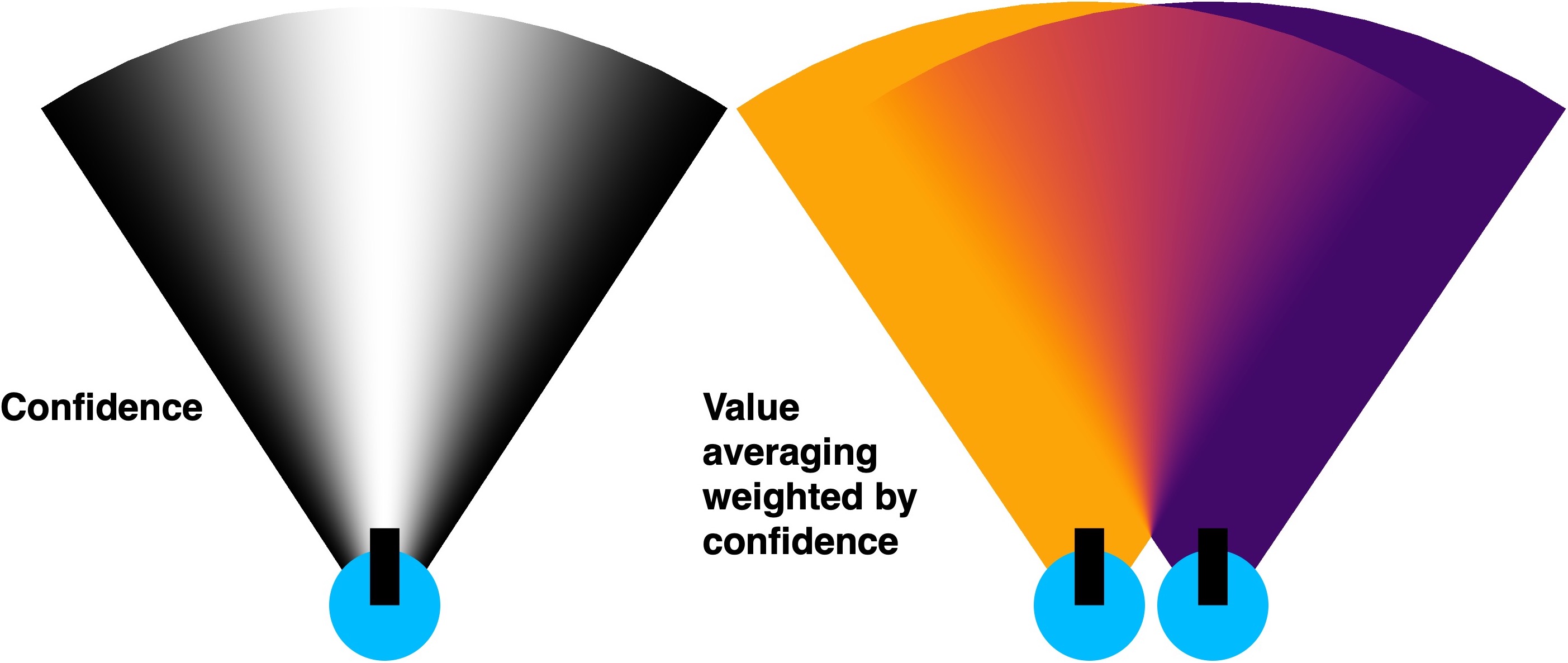}}
    \caption{
        \textit{Left:} Visualization of how the confidence score of a pixel within the robot's FOV is determined based on its location relative to the optical axis.
        \textit{Right:} The confidence scores are used when the robot's current FOV overlaps with the previously seen area; the new semantic values within this region become an average of the previous and current values, weighted by their confidence scores.
    }
    \vspace{-0.2cm}
    \label{fig:confidence}
\end{figure}

\begin{figure*}[t!]
\centering
    \vskip5pt
    \includegraphics[width=1.0\textwidth]{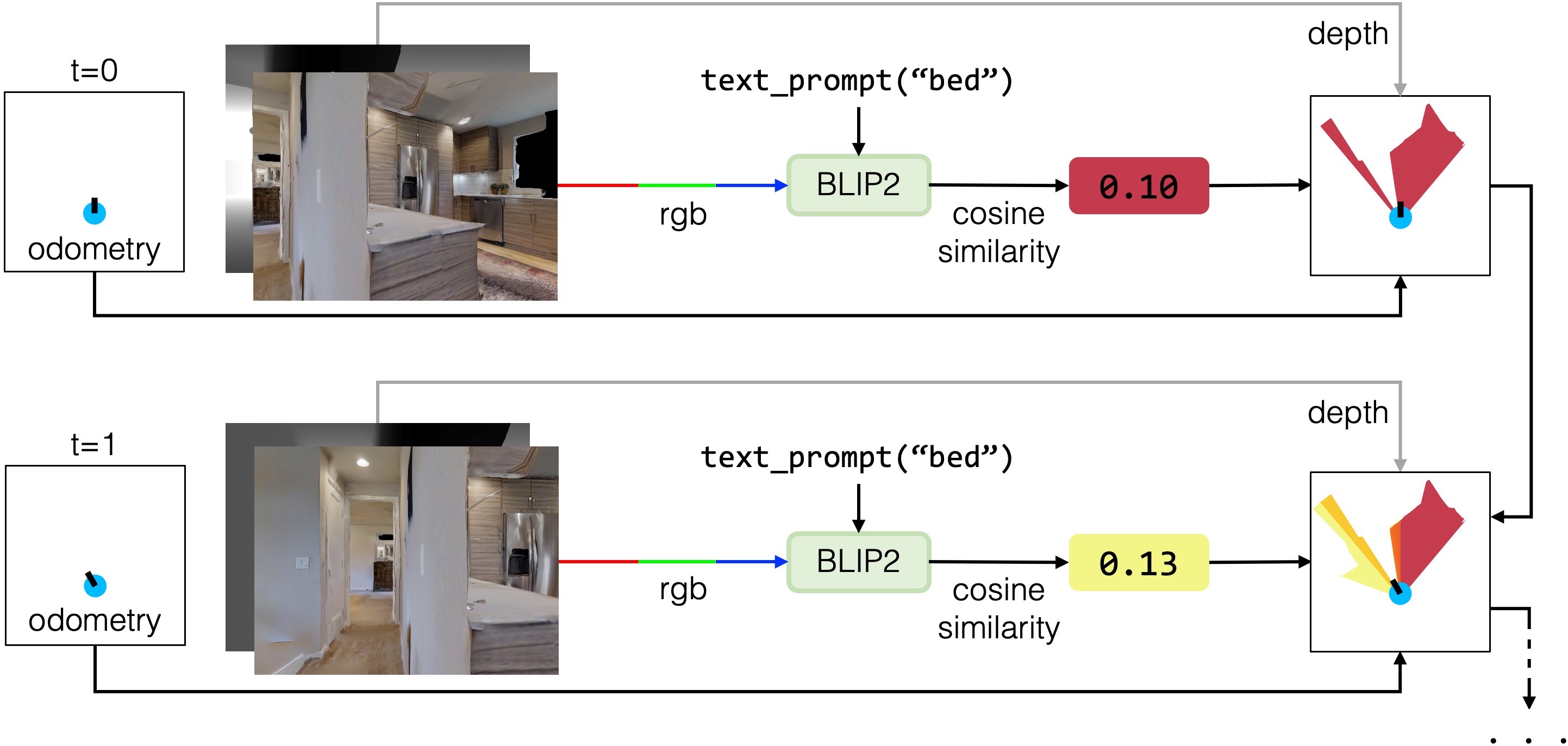}
    \caption{
        \approachname\ iteratively constructs value maps for target-driven navigation by using \mbox{BLIP-2} to compute the cosine similarity between a text prompt incorporating the target object and an RGB image taken from the robot's current pose. These semantic value scores are projected onto a top-down 2D pixel grid in the shape of the camera's FOV and exclude regions occluded by obstacles captured in the depth image.
    }
    \label{fig:value_map}
    \vspace{-0.2cm}
\end{figure*}

At the core of our approach is a value map, a 2D grid similar to the frontier map. 
This map assigns a value to each pixel within the explored area, quantifying its semantic relevance in locating the target object.
The value map is used to evaluate each frontier, and the frontier with the highest value is chosen as the next one to explore. 
Similar to the frontier map, the value map uses depth and odometry observations to build a top-down map iteratively. 
However, the value map differs in that it has two channels representing semantic value scores and confidence scores.

Similar to how humans derive semantic cues directly from visual observations (\eg, lighting, room type, room size, navigability to other rooms), rather than attempting to first represent what is currently visible to the robot with text (\eg, using detected object bounding boxes like \cite{dorbala2023can,zhou2023esc,chen2023not}), we use a pre-trained \mbox{BLIP-2}~\cite{li2023blip} vision-language model to compute a cosine similarity score directly from the robot's current RGB observation and a text prompt containing the target object.
\mbox{BLIP-2} adapts CLIP~\cite{radford2021learning} to achieve state-of-the-art results for image-to-text retrieval, which relies on accurately measuring how well a text prompt is represented by a given image.
When used for image-to-text retrieval, \mbox{BLIP-2} outputs a cosine score given an input RGB image and text prompt, where higher values indicate higher accuracy.
We use a text prompt to measure how valuable the area represented by the current RGB image is for finding the target object (\texttt{"\targetprompt"}).
These scores are then projected onto their own channel of the top-down value map.

The confidence channel aims to determine how a pixel's value in the semantic value channel should be updated if it has a value assigned from a previous time step and is within the robot's field-of-view (FOV) at the current time step.
It does not affect a pixel's semantic value score if that pixel was not seen until the current time step.
The confidence score of a pixel within the robot's FOV depends on its location relative to the optical axis.
Pixels along the optical axis have a full confidence of 1, while those at the left and right edges have a confidence of 0.
Specifically, we set the confidence of a pixel as $cos^2(\theta / (\theta_{fov}/2) * \pi/2)$, where $\theta$ is the angle between the pixel and the optical axis, and $\theta_{fov}$ is the horizontal FOV of the robot's camera.

When the robot moves to a new position where its FOV overlaps with a previously seen region, the semantic value and confidence scores for each pixel in that region are both updated with new scores.
Each of these pixels' new semantic value score, $v^{new}_{i,j}$, is computed by averaging its current and previous value scores, weighted by $c^{curr}_{i,j}$ and $c^{prev}_{i,j}$, its current and previous confidence scores: $v^{new}_{i,j}=(c^{curr}_{i,j}v^{curr}_{i,j}+c^{prev}_{i,j}v^{prev}_{i,j})/ (c^{curr}_{i,j}+c^{prev}_{i,j})$.
Its new confidence score is also updated using a weighted average that is biased towards the higher confidence value:
$c^{new}_{i,j}=((c^{curr}_{i,j})^2+(c^{prev}_{i,j})^2)/ (c^{curr}_{i,j}+c^{prev}_{i,j})$.
The confidence channel and its role in updating previously seen values is visualized in Fig.~\ref{fig:confidence}.

The map updating procedure is summarized as follows:
1) a cone-shaped mask depicting the camera FOV in a top-down manner at the camera's current pose is created, where pixels closer to the optical axis of the camera have a higher confidence score;
2) using the depth image, the mask is updated to exclude areas of the FOV obstructed by obstacles;
3) using \mbox{BLIP-2}, cosine similarity scores are computed between the current RGB image and the text prompt to update the semantic value channel within the masked portion of the value map;
4) previous semantic value and confidence scores within the masked portion of the value map are updated using weighted averaging using the confidence scores.
The full process is depicted in Fig.~\ref{fig:value_map}.

\subsection{Object detection}
To determine whether or not a target object instance is currently visible to the robot, we use pre-trained object detectors that infer bounding boxes with semantic labels.
Specifically, we utilize YOLOv7~\cite{wang2022yolov7} for target objects that fall within the COCO~\cite{lin2014microsoft} classes, and Grounding-DINO~\cite{liu2023grounding} for all other categories.
\mbox{Grounding-DINO} is an open-vocabulary object detector capable of detecting arbitrary objects based on language inputs.
We do this because we find that YOLOv7 is better at detecting the objects within the COCO categories.
If a target object instance is detected, we use \mbox{Mobile-SAM}~\cite{mobile_sam} to extract its contour using the RGB image and the detected bounding box.
The contour is then used with the depth image to determine the point on the object that is closest to the robot's current position, which is then used as the goal waypoint to navigate to.
Once the robot's distance to this point falls below the success radius, \stopac is called. 

\subsection{Waypoint navigation}

After initialization, the robot is always provided with either a frontier waypoint or a target object waypoint to navigate towards, depending on whether a target object has been detected yet.
To determine the action at each step for reaching the current waypoint, we employ a Point Goal Navigation (PointNav)~\cite{anderson_arxiv18} policy. 
To determine the action at each step for reaching the current waypoint,
we use Variable Experience Rollout (VER)~\cite{wijmans2022ver}, a distributed deep reinforcement learning algorithm, to train a PointNav~\cite{anderson_arxiv18} policy.
PointNav is a task that challenges the robot to navigate to a designated waypoint (2D coordinate) solely relying on visual observations and odometry.
We trained our PointNav policy using scenes from the training split of the HM3D dataset~\cite{ramakrishnan2021hm3d}, using the same hyperparameters as those described in \cite{wijmans2022ver}, with 4 GPUs (64 workers each), and train the policy for 2.5 billion steps (around 7 days).
Unlike ObjectNav, PointNav does not necessitate a semantic understanding of the environment for efficient and successful completion and can be accomplished using only geometric understandings.
Our PointNav policy exclusively uses the egocentric depth image and the robot's relative distance and heading towards the desired goal point as input (see Fig.~\ref{fig:overview}).
It does not use RGB images.

It is entirely feasible to replace this PointNav policy with any alternative method capable of guiding the robot to a visually observed waypoint, be it a frontier or a detected target object.
For example, we do not use a PointNav policy for our real-world demonstrations.
Our preference for the PointNav policy stems from its speed and ease-of-use;
it alleviates several concerns associated with traditional mapping-based approaches, particularly when the waypoint resides outside the navigable area (\eg, when the waypoint is on a target object, which itself may also be on a different obstacle), as navigability of the goal does not affect the policy or its observations.
\section{Experimental Setup}

\xhdr{Datasets.}
We evaluate our approach using the Habitat~\cite{habitat19iccv} simulator on the validation splits of three different datasets of 3D scans of real-world environments; Gibson~\cite{xiazamirhe2018gibsonenv}, HM3D~\cite{ramakrishnan2021hm3d}, and MP3D~\cite{Matterport3D}. We use the ObjectNav validation split for Gibson developed in SemExp~\cite{chaplot2020object} which contains 1000 episodes across 5 scenes.
HM3D's validation split contains 2000 episodes across 20 scenes and 6 object categories. MP3D's validation split contains 2195 episodes across 11 scenes and 21 object categories.

\xhdr{Metrics.}
For all approaches, we report success rate (SR) and Success weighted by inverse Path Length (SPL)~\cite{anderson_arxiv18}.
SPL scores the efficiency of an agent's path by comparing it to the length of the shortest path from the start position to the closest instance of the target object category.
It is zero if the agent did not succeed; otherwise, it is the shortest path length divided by the agent's path length (larger is better).

\xhdr{Baselines.}
We evaluate \approachname\ by comparing it to several state-of-the-art (SOTA) techniques for zero-shot object navigation: CLIP on Wheels (CoW)~\cite{gadre2022cow}, ESC~\cite{zhou2023esc}, SemUtil~\cite{chen2023not}, and ZSON~\cite{majumdar2022zson}.
ZSON is an open vocabulary method that uses CLIP to transfer a method for a different navigation task (ImageNav) zero-shot to the ObjectNav task.
It is trained on ImageNav in which images are used as goals instead of object names; at test-time, text-embeddings of object category names are used as goals instead.
CoW explores the closest frontier until the target object is detected using either CLIP features or an open-vocabulary object detector, and then navigates directly to the detected target.
Similarly to our approach, ESC and SemUtil both perform semantic frontier-based exploration, but frontiers are evaluated using nearby object detections that are converted to text and evaluated using a text-only model (\eg, an LLM, BERT) that also considers the target object category.

In addition to the above zero-shot SOTA methods, we also include a comparison with supervised methods: PONI~\cite{ramakrishnan2022poni}, PIRLNav~\cite{ramrakhya2023pirlnav}, RegQLearn~\cite{gireesh22objQlearn}, and SemExp~\cite{chaplot2020object}. SemExp and PONI both build maps during navigation and train task-specific policies to perform semantic inference to predict likely target locations to explore. PIRLNav is an end-to-end policy trained with behavioral cloning on 77\textit{k} human demonstrations and fine-tuned with online deep reinforcement learning. RegQLearn is an end-to-end policy trained only with deep reinforcement learning.

\section{Results}
\begin{table}[t]
    \centering
    \setlength{\abovecaptionskip}{8pt}
    \setlength{\belowcaptionskip}{8pt}
    \caption{Zero-shot Object Navigation results on Gibson~\cite{xiazamirhe2018gibsonenv}, HM3D~\cite{ramakrishnan2021hm3d}, and MP3D~\cite{Matterport3D} benchmarks. Our method outperforms previous zero-shot methods and performs competitively against methods directly trained on the Object Navigation task. 
    }
    \vskip 0.1in
    \resizebox{\columnwidth}{!}{
    \begin{tabular}{@{} l@{\hspace{1pt}}c@{\hspace{4pt}}c@{\hspace{4pt}}c@{\hspace{4pt}}c@{\hspace{4pt}}c@{\hspace{4pt}}c@{\hspace{4pt}}c@{\hspace{4pt}}c @{}}
        \toprule
        \multirow{2}*{\textbf{Approach}} & \multirow{2}*{\textbf{Semantic Nav}} &
        \multicolumn{2}{c}{\textbf{Gibson}} & \multicolumn{2}{c}{\textbf{HM3D}} & \multicolumn{2}{c}{\textbf{MP3D}} \\
        \cmidrule(lr){3-4}\cmidrule(lr){5-6}\cmidrule(lr){7-8}
        & \textbf{Training} & SPL$\uparrow$ & SR$\uparrow$ & SPL$\uparrow$ & SR$\uparrow$ & SPL$\uparrow$ & SR$\uparrow$ \\
        \midrule
        $\mathrm{PONI}$~\cite{ramakrishnan2022poni} & ObjectNav & 41.0 & 73.6 & - & - & 12.1 & 31.8 \\
        $\mathrm{PIRLNav}$~\cite{ramrakhya2023pirlnav}  & ObjectNav & - & - & 27.1 & \textbf{64.1} & - & - \\ 
        $\mathrm{RegQLearn}$~\cite{gireesh22objQlearn}  & ObjectNav & 31.3 & 63.7 & - & - & - & - \\ 
        $\mathrm{SemExp}$~\cite{chaplot2020object}  & ObjectNav & 33.9 & 65.7 & - & - & - & - \\ 
        \midrule
        $\mathrm{ZSON}$~\cite{majumdar2022zson} & ImageNav & - & - & 12.6 & 25.5 & 4.8 & 15.3 \\
        $\mathrm{CoW}$~\cite{gadre2022cow}  & None & - & - & - & - & 3.7 & 7.4 \\
        $\mathrm{ESC}$~\cite{zhou2023esc}  & None & -  & - & 22.3  & 39.2 & 14.2 & 28.7 \\
        $\mathrm{SemUtil}$~\cite{chen2023not}  & None & 40.5 & 69.3 & - & - & - & - \\
        $\mathrm{\approachname}$ (Ours) & None & \textbf{52.2} & \textbf{84.0} & \textbf{30.4} & 52.5 & \textbf{17.5} & \textbf{36.4} \\
        \bottomrule
        \bottomrule
    \end{tabular}
    }
    \label{tab:main_results}
\end{table}

In this section, we aim to address the following questions:
\begin{enumerate}
    \item How well can \approachname\ perform ObjectNav in various datasets in comparison to other trained or zero-shot methods?
    \item How do different methods of fusing current and previously seen values affect the performance of \approachname?
    \item Can \approachname\ be deployed successfully in the real world?
\end{enumerate}

\subsection{Benchmark results}
The performance of \approachname\ in comparison to other methods on the Gibson, HM3D, and MP3D datasets is summarized in Table \ref{tab:main_results}.
Unfilled cells indicate that the cited work did not evaluate their approach on the corresponding dataset (SemExp only evaluated on a subset of MP3D episodes using COCO classes).
\approachname\ significantly outperforms all zero-shot methods across all benchmarks, with an increase of +11.7\% SPL and +14.7\% success in Gibson compared to SemUtil; +8.1\% SPL and +13.3\% success in HM3D compared to ESC; and +3.3\% SPL and +7.7\% success in MP3D compared to ESC.

In the Gibson and MP3D datasets, \approachname\ even surpasses methods that were trained directly within those datasets for ObjectNav, achieving +19.2\% SPL and +19.0\% success in Gibson compared to SemExp, and +5.4\% SPL and +4.6\% success in MP3D compared to PONI.
This establishes new state-of-the-art metrics for these datasets. 

In the HM3D dataset, \approachname\ only falls short of PIRLNav in terms of success (-11.6\% success, but +3.3\% SPL), which was trained on 77\textit{k} human demonstrations collected within the train split of the HM3D dataset, while our method is entirely zero-shot.

Our superior performance within the Gibson dataset can be attributed to the lack of episodes that require the robot to navigate stairs in order to reach a target object, a scenario present in both HM3D and MP3D.
The only non-zero-shot approach to outperform \approachname\ in success rate, PIRLNav, can traverse up or down stairs to find objects.
However, \approachname\ currently only supports single-floor episodes due to the lack of a $z$ coordinate in the odometry observation provided by the simulator, which complicates the resetting of top-down frontier and value maps upon changing floors.
Consequently, we fail 14.6\% and 9.6\% of the HM3D and MP3D episodes (respectively) because they require traversing stairs to encounter the target object.

We also attribute the higher performance on the Gibson and HM3D datasets compared to the MP3D dataset to the quality of the 3D scans.
The MP3D dataset has significantly lower visual fidelity~\cite{ramakrishnan2021hm3d} than the HM3D dataset, while the scenes from the Gibson dataset were manually repaired and verified to be free of holes and artifacts~\cite{xiazamirhe2018gibsonenv}. 

\begin{table}[t]
    \centering
    \setlength{\abovecaptionskip}{8pt}
    \setlength{\belowcaptionskip}{8pt}
    \caption{Comparisons of performance with different value update methods used with \approachname.
    }
    \vskip 0.1in
    \resizebox{\columnwidth}{!}{
    \begin{tabular}{@{}lc@{\hspace{4pt}}cc@{\hspace{4pt}}cc@{\hspace{4pt}}c}
        \toprule
        \multirow{2}*{\textbf{Value update}} & \multicolumn{2}{c}{\textbf{Gibson}} & \multicolumn{2}{c}{\textbf{HM3D}} & \multicolumn{2}{c}{\textbf{MP3D}} \\
        \cmidrule(lr){2-3}\cmidrule(lr){4-5}\cmidrule(lr){6-7}
        \textbf{method} & SPL$\uparrow$ & SR$\uparrow$ & SPL$\uparrow$ & SR$\uparrow$ & SPL$\uparrow$ & SR$\uparrow$ \\
        \midrule
        Replacement & 48.0 & 76.1 & 26.5  & 44.5 & 16.6 & 31.7 \\
        Unweighted avg. & 50.9 & 83.0 & 30.0 & 51.8 & 17.1 & 35.0 \\
        Weighted avg. & \textbf{52.2} & \textbf{84.0} & \textbf{30.4} & \textbf{52.5} & \textbf{17.5} & \textbf{36.4} \\
        \bottomrule
    \end{tabular}
    }
    \label{tab:ablations}
    \vspace{-2ex}
\end{table}

\subsection{Ablations}
When an area that was already seen previously is encountered again, its representation in the value map used by \approachname\ is updated using a combination of its previous and current values.
We explore different methods of combining these values and their impact on the performance of \approachname\ in Table \ref{tab:ablations}. 
In the \textit{Replacement} method, the previous value is disregarded and the new value simply overwrites it.
The \textit{Unweighted avg.} method calculates the new value as the average of the previous and current values.
Our full approach, \textit{Weighted avg.}, uses confidence scores to weight the average between the previous and new values, taking into account the parts of the FOV the area occupied when it was previously observed and at the current timestep. 
Our findings indicate that the \textit{Weighted avg.} method consistently enhances performance compared to the other two methods across all three datasets.

\subsection{Real-world deployment}
We deploy \approachname\ on a Spot robot from Boston Dynamics (BD) to demonstrate \approachname\ successfully navigating to objects in the real world.
For waypoint navigation, we utilize the BD API instead of a PointNav policy, which can guide the robot towards a specified waypoint, provided the path is relatively free of obstacles.
The camera within the robot's gripper is used to detect objects and feed inputs to \mbox{BLIP-2}. 
However, due to its limited range in depth sensing, we employ the ZoeDepth depth estimation model~\cite{zoedepth} to generate a depth image and approximate a waypoint for the detected target object.
We rely on the depth cameras on the body of Spot to detect obstacles for frontier generation.
All models used, including \mbox{BLIP-2}, GroundingDINO, MobileSAM, and ZoeDepth, were loaded and executed in real-time on a laptop equipped with an RTX 4090 MaxQ Mobile GPU with 16 GB of VRAM. Videos can be viewed at \projecturl.
\section{Conclusion}
This paper presents \approachname, a zero-shot framework for ObjectNav in novel environments.
Our key innovation is spatially grounding joint vision-language-based semantic reasoning with pre-trained models in a new approach to frontier waypoint selection in order to perform target-driven navigation in novel environments.
\approachname\ uses a semantic prompt-based method to infer which frontier to navigate to using a pre-trained vision-language model, detects visible target objects with a pre-trained object detector, and navigates to these frontiers and target objects using a pre-trained policy. 
This modular nature of our method allows different components to be swapped in as improved models become available.
Experiments in simulated 3D home environments demonstrate that \approachname\ achieves state-of-the-art zero-shot navigation performance on the Object Goal Navigation benchmark. Demonstrations in an office building on a Spot Arm platform prove the viability of our method in real-world scenarios.

\approachname\ has a number of limitations that could be addressed by future work. %
First, we assume target objects will be easily visible in the scene from the default height of the robot camera. Future work could investigate policies to increase interaction with the environment during the search process, including actively directing the robot camera to look in promising locations and using manipulation to execute search actions such as looking inside closed drawers. Furthermore, we build a map of semantic information and occupancy information with our frontier and value map, but the value map, which contains semantic information, provides only task-specific semantic information. So, we cannot leverage this map in sequentially executed semantic navigation tasks to different objects or in executing other navigation tasks requiring targets specified by language, such as vision-language navigation.
Future work could explore alternate prompt formulations, value map designs, and methods of tracking semantic information during task execution to enable long-horizon planning and multitask execution.

Overall, our results highlight the promise of leveraging foundation models like \mbox{BLIP-2} in a zero-shot manner within robotic systems to provide spatially grounded joint vision-language-based semantic reasoning without task-specific training.

\section{Acknowledgements}
\scriptsize{
    The Georgia Tech effort was supported in part by ONR YIPs, ARO PECASE, and Korea Evaluation Institute of Industrial Technology (KEIT) funded by the Korea Government (MOTIE) under Grant No.20018216, Development of mobile intelligence SW for autonomous navigation of legged robots in dynamic and atypical environments for real application. The views and conclusions are those of the authors and should not be interpreted as representing the U.S. Government, or any sponsor.
}

\addtolength{\textheight}{-5cm}   %

\bibliographystyle{IEEEtran}
\bibliography{references}

\end{document}